\documentclass[10pt,twocolumn,letterpaper]{article}

\usepackage[pagenumbers]{cvpr} % To force page numbers, e.g. for an arXiv version

\usepackage{graphicx}
\usepackage{amsmath}
\usepackage{amssymb}
\usepackage{booktabs}
\usepackage[table]{xcolor}
\usepackage{tabulary,multirow,overpic,xcolor}
\definecolor{baselinecolor}{gray}{.92}

\newcommand{\sref}[1]{\S\ref{#1}}

\usepackage{tabulary}

\definecolor{defaultcolor}{gray}{.92}
\definecolor{citecolor}{HTML}{0071bc}
\definecolor{darkgreen}{rgb}{0.0, 0.2, 0.13}

\newcommand{\cbr}[1]{\left\{#1\right\}}

\definecolor{green}{HTML}{39b54a}  % green
\definecolor{red}{HTML}{ea4335}  % red
\newcommand{\hlg}[1]{\textcolor{chameleon3}{#1}}

\def\x{$\times$}

\newcommand{\save}[2]{#1 \hlg{$\downarrow$\scriptsize{#2}}
}

\newcommand{\increase}[2]{
#1 \hlg{$\uparrow$\scriptsize{#2}}
}

\newcommand{\app}{\raise.17ex\hbox{$\scriptstyle\sim$}}
\newcommand{\specialcell}[2][c]{%
  \begin{tabular}[#1]{@{}c@{}}#2\end{tabular}}
\def\x{$\times$}

\definecolor{carmine}{rgb}{0.59, 0.0, 0.09}

\newcolumntype{x}[1]{>{\centering\arraybackslash}p{#1pt}}
\newcolumntype{y}[1]{>{\raggedright\arraybackslash}p{#1pt}}
\newcolumntype{z}[1]{>{\raggedleft\arraybackslash}p{#1pt}}
\newlength\savewidth\newcommand\shline{\noalign{\global\savewidth\arrayrulewidth
		\global\arrayrulewidth 1pt}\hline\noalign{\global\arrayrulewidth\savewidth}}
\newcommand{\tablestyle}[2]{\setlength{\tabcolsep}{#1}\renewcommand{\arraystretch}{#2}\centering\footnotesize}

\makeatletter
\@namedef{ver@everyshi.sty}{}
\makeatother

\usepackage{tikz,pgfplots}
\usepackage{tikz,siunitx}
\usepackage{tango}

\definecolor{demphcolor1}{gray}{.5}

\newcommand{\demph}[1]{\textcolor{demphcolor1}{#1}}

\definecolor{xycolor}{RGB}{60, 120, 216}
\definecolor{xycolor}{HTML}{0071bc}
\newcommand{\xycolor}[1]{\textcolor{xycolor}{#1}}
\definecolor{wcolor}{RGB}{103, 78, 167}
\newcommand{\wcolor}[1]{\textcolor{wcolor}{#1}}
\definecolor{dcolor}{RGB}{166, 77,21}

\definecolor{gcolor}{RGB}{204, 102, 153}

\definecolor{tcolor}{RGB}{80, 200, 180}
\newcommand{\tcolor}[1]{\textcolor{citecolor}{#1}}
\definecolor{eicolor}{RGB}{153, 51, 102}
\newcommand{\eicolor}[1]{\textcolor{eicolor}{#1}}

\newcommand{\outsizesRawAdapt}[4]{\multirow{#4}{*}{\(\begin{array}{c}  \text{#2\x #3}\\[-.1em]  \end{array}\)}}

\newcommand{\outsizesRawDAdapt}[5]{\multirow{#5}{*}{\(\begin{array}{c}  \text{#1\x#3\x#4}\\[-.1em]  \end{array}\)}}

\newcommand{\blockatt}[3]{\multirow{2}{*}{\(\left[\begin{array}{c}\eicolor{\mathbf{F}}:\text{ MHA(\wcolor{#1})}\\[-.1em] \eicolor{\mathbf{G}}:\text{MLP(\wcolor{#2})}\end{array}\right]\)$\times$#3}
}

\newcommand{\blockatta}[3]{\multirow{2}{*}{\(\left[\begin{array}{c}\eicolor{\mathbf{F}}: \text{MHPA(\wcolor{#1})}\\[-.1em] \eicolor{\mathbf{G}}:\text{MLP(\wcolor{#2})}\end{array}\right]\)$\times$#3}
}

\newcommand{\outputstack}[3]{\multirow{2}{*}{\(\left[\begin{array}{c}Y_1: \text{#1\x#2\x#3}\\[-.1em] Y_2: \text{#1\x#2\x#3}\end{array}\right]\)}
}
\newcommand{\blockattfus}[3]{\multirow{3}{*}{\(\left[\begin{array}{c} \text{\eicolor{FUSION}(\wcolor{#1})}\\[-.1em] 
\text{MHPA(\wcolor{#1})}\\[-.1em] 
\text{MLP(\wcolor{#2})}\\
\end{array}
\right]\)$\times$#3}
}

\definecolor{baselinecolor}{gray}{.92}

\definecolor{citecolor}{RGB}{34,139,34}
\definecolor{citecolor2}{HTML}{0071bc}
\definecolor{Graylight}{gray}{0.9}
\definecolor{lightred}{RGB}{241,140,142}
\usepackage[pagebackref=true,breaklinks=true,letterpaper=true,colorlinks,
citecolor=citecolor2,bookmarks=false]{hyperref}

\usepackage[capitalize]{cleveref}
\usepackage{ragged2e} 
\crefname{section}{Sec.}{Secs.}
\Crefname{section}{Section}{Sections}
\Crefname{table}{Table}{Tables}
\crefname{table}{Tab.}{Tabs.}

\def\cvprPaperID{***} % *** Enter the CVPR Paper ID here
\def\confName{CVPR}
\def\confYear{2022}

\def\x{$\times$}

\newcommand{\expnum}[2]{{#1}\mathrm{e}{#2}}

\definecolor{xycolor}{RGB}{60, 120, 216}
\definecolor{xycolor}{HTML}{0071bc}
\definecolor{wcolor}{RGB}{103, 78, 167}
\definecolor{dcolor}{RGB}{166, 77,21}
\definecolor{gcolor}{RGB}{204, 102, 153}
\definecolor{tcolor}{RGB}{80, 200, 180}
\definecolor{eicolor}{RGB}{153, 51, 102}

\def\x{$\times$}
 
\pgfplotsset{compat = 1.3,
	legend style={font=\scriptsize},
	legend cell align={left},
	legend style={cells={align=left}, draw=black!20},
	grid=both,
	grid style={dotted},
	tick style={draw=none},
	enlarge x limits=false,
	enlarge y limits=false,
	axis line style={draw=black!100},
	axis lines=left,
}

\pgfplotscreateplotcyclelist{mycolormarklist}{%
	chameleon3,mark=diamond*,mark options={solid, fill=chameleon3}\\
	skyblue2,mark=pentagon*,mark options={solid}\\
	scarletred3,mark=triangle*,mark options={solid}\\
	plum1,mark=square*,mark options={solid}\\
	aluminium5,mark=otimes*,mark options={solid}\\
	orange1,mark=square*,mark options={solid}\\
	skyblue3\\
	chocolate2\\
	maroon\\
	butter3\\
}
\pgfplotscreateplotcyclelist{mystylelist}{%
	densely dashdotted\\
	solid\\
	dotted\\
	densely dashed\\
	dashed\\
	dashdotted\\
	densely dotted\\
	densely dash dot dot\\
	dotted\\
	dash dot dot\\
}

\usepackage[capitalize]{cleveref}
\crefname{section}{Sec.}{Secs.}
\Crefname{section}{Section}{Sections}
\Crefname{table}{Table}{Tables}
\crefname{table}{Tab.}{Tabs.}

\def\dashedrule#1#2#3{{%
\dimen1=#2 \divide\dimen1 by 2
\def\@ruledash{%
\rule{\dimen1}{0pt}%
\rule[0.5ex]{#1}{0.4pt}%
\rule{\dimen1}{0pt}}%
\count1=0
\loop%
\ifnum\count1<#3%
\advance\count1 by 1%
\@ruledash%
\repeat}}

\def\mediumdashes{\dashedrule{.3em}{.2em}{4}}

\begin{document}
	
	%%%%%%%%% TITLE - PLEASE UPDATE
	\title{ \vspace{-.7em} Reversible Vision Transformers  \vspace{-.7em}}
	\author{
	Karttikeya Mangalam\textsuperscript{*, 2} \qquad 
	Haoqi Fan\textsuperscript{1} \qquad 
	Yanghao Li\textsuperscript{1} \qquad \\ 
    Chao-Yuan Wu\textsuperscript{1} \qquad
	Bo Xiong\textsuperscript{ 1}\qquad
	Christoph Feichtenhofer\textsuperscript{ *, 1} \qquad
	Jitendra Malik\textsuperscript{ 1, 2} \qquad \\ 
	\small   \\
	\textsuperscript{1}Meta AI, FAIR \qquad \qquad \textsuperscript{2}UC Berkeley 
  % \vspace{.5em} 
}
	\maketitle
	
	\begin{abstract}
We present Reversible Vision Transformers, a memory efficient architecture design for visual recognition. By decoupling the GPU memory requirement from the depth of the model, Reversible Vision Transformers enable scaling up architectures with efficient memory usage. We adapt two popular models, namely Vision Transformer and Multiscale Vision Transformers, to reversible variants and benchmark extensively across both model sizes and tasks of image classification, object detection and video classification. Reversible Vision Transformers achieve a reduced memory footprint of up to \textbf{15.5}\x~at roughly identical model complexity, parameters and accuracy, demonstrating the promise of reversible vision transformers as an efficient backbone for hardware resource limited training regimes. Finally, we find that the additional computational burden of recomputing activations is more than overcome for deeper models, where throughput can increase up to $\textbf{2.3}$\x~over their non-reversible counterparts. Full code and trained models are available at \url{https://github.com/facebookresearch/slowfast}. A simpler, easy to understand and modify version is also available at \url{https://github.com/karttikeya/minREV}. 
	\end{abstract}
 
\renewcommand*{\thefootnote}{\fnsymbol{footnote}}
\setcounter{footnote}{1}
\footnotetext{Equal technical contribution.}
\renewcommand*{\thefootnote}{\arabic{footnote}}
\setcounter{footnote}{0}

	\section{Introduction}
	\label{sec:intro}
	The deep learning revolution in computer vision has rested on the bedrock of high performance hardware accelerators. Fueled by special purpose AI accelerators, the compute requirements for state-of-the-art models are growing exponentially. 
	However, compute is only half the story. The other, and often overlooked half, is memory bandwidth bottleneck, which has been difficult to proportionally scale as compared to peak accelerator FLOPs~\cite{patterson2004latency}.
	In particular, the peak accelerator FLOPs have been increasing at a rate of $\app$3.1\x\ every 2 years~\cite{gholami2021, sun2019summarizing}. However, peak bandwidth only scales at a rate of $\app$1.4\x\ every 2 years.
	This disparity is exacerbated in transformers, which have been doubling in required compute roughly every three months for the past three years, resulting in a so-called memory wall~\cite{gholami2021} where both the overall model performance as well as the training speed have become tightly memory-bound~\cite{ivanov2020data}. 
	
    As such, for bandwidth bound models, trading compute for memory through re-computation could actually be more efficient than using work-optimal algorithms~\cite{williams2008auto, williams2009roofline}.
    In the case of training neural network models, this can be achieved by re-computing activations instead of storing and then loading them from DRAM~\cite{horowitz20141}. 
	Besides training speed, scaling vision transformers in depth naturally hits the GPU memory capacity, especially in memory starved regimes such as video recognition where state-of-the-art models are often limited to batch size $1$ due to high memory footprint of intermediate activations. 
		
	%%%%%%%%%%%%%%%%%%%%%%%%%%%%%%%%%%%%%%%%%%%%%%%%%%%%%%%%%%%%%%%%%%%%%%%%%%%%%%%
	\begin{figure}[t]
		\centering
  \vspace{-10pt}
		\includegraphics[width = \linewidth]{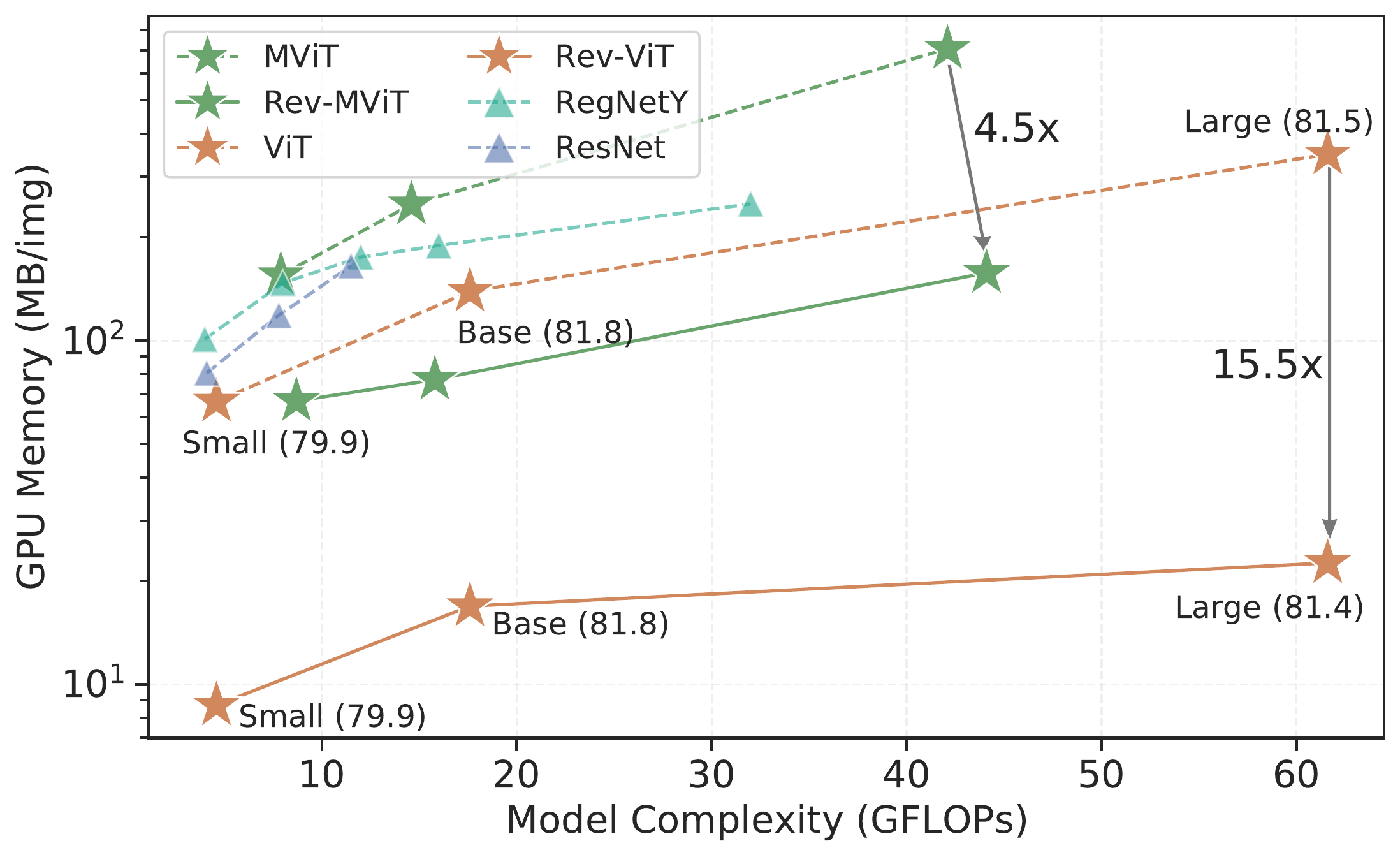}
\caption{\textbf{Reversible Vision Transformers} are more memory-efficient, yet powerful \textit{reversible counterparts} of state-of-the-art Vision Transformer (ViT)~\cite{ViT} and Multiscale Vision Transformer (MViT)~\cite{MViT} architectures with varying model complexity. Numbers in parentheses denote top-1 ImageNet performance. ResNet~\cite{He2016} and RegNet~\cite{ilija_2020} are only shown for reference. For detailed discussion please refer to \sref{sec:results:IN}.
}
		\label{fig:teaser}
  \vspace{-10pt}
  	\end{figure}
	%%%%%%%%%%%%%%%%%%%%%%%%%%%%%%%%%%%%%%%%%%%%%%%%%%%%%%%%%%%%%%%%%%%%%%%%%%%%%%%
	We propose Reversible Vision Transformers, a family of expressive visual recognition architectures with very favorable activation memory footprints (Figure \ref{fig:teaser}) compared to their non-reversible variants.
	By trading-off GPU activation caching with efficient on-the-fly activation re-computation, reversible vision transformers effectively \textit{decouple} the activation \textit{memory} growth from the \textit{depth} of the model. 

	While the natural language processing community has performed some early exploration of reversible transformers for machine translation~\cite{kitaev2020reformer}, these techniques focus on \textit{longer sequence lengths} rather than depth.

	Our experiments show that a straightforward adaptation of vision transformers to reversible architectures \textit{fails} to scale for \textit{deeper} models because of training convergence instabilities due to internal sub-block residual connections.

	In this work, we reconfigure the residual paths in Vision Transformers  (ViT)~\cite{ViT} and Multiscale Vision Transformers (MViT)~\cite{MViT} to overcome this issue. We further find that reversible structures have stronger inherent regularization and therefore, we use a lighter augmentation recipe (repeated augmentation, augmentation magnitude and stochastic depth) and lateral connections between residual blocks.
	
	We benchmark extensively across image recognition tasks such as image classification and object detection as well as video classification, across all of which, reversible vision transformers have competitive performance to their non-reversible counterparts suffering negligible to no performance decay.
	Moreover, reversible models have extremely favorable per-image memory footprint, saving $\textbf{15.5}$\x~on the ViT-Large model and $\textbf{4.5}$\x~on the MViT-Large model with reversible training.
	
    In summary, our contributions are three-fold. 
    
    (\textit{i}) We propose Reversible Vision Transformer (\textbf{\mbox{Rev-ViT}}) and Reversible Multiscale Vision Transformers (\textbf{\mbox{Rev-MViT}}), memory efficient reversible adaptations of state-of-the-art visual recognition backbones. 
    
    (\textit{ii}) We observe reversible transformers to have a stronger inherent regularization than vanilla networks. Hence, we develop new training recipes by adapting the original recipes with different repeated augmentations, augmentation magnitudes and drop path rate to match the performance of their non-reversible counterparts. 
    
    (\textit{iii}) We benchmark our models across several tasks: image classification, object detection and action recognition, across accuracy, memory, maximum training batch size and model complexity. In particular, at matched complexity (FLOPs/parameters) and final accuracy, \mbox{Rev-ViT-B} and  Rev-ViT-L train with per image memory footprints that are 8.2\x\ and 15.5\x\ lighter than ViT-B and ViT-L respectively. Further, we show how \textit{deep} reversible networks can achieve up to 2-4\x~{throughput} than their vanilla counterparts. 
	
	\section{Related Work}
	\label{sec:related}
	\paragraph{Transformers} are a popular network structure that were first proposed for natural language applications~\cite{vaswani2017attention} and now are widely used in all areas of deep learning such as Reinforcement Learning~\cite{chen2021decision}, Speech~\cite{li2019neural}, Music~\cite{huang2018music}, multi-modal learning~\cite{jaegle2021perceiver} and recently, in traditional vision tasks~\cite{ViT} as well. Since their introduction, Vision Transformers have experienced enthusiastic adoption and have been applied to several visual recognition tasks~\cite{ViT,deit,touvron2021going} using priors such as multi-scale feature hierarchies~\cite{MViT,Graham_2021_ICCV,Swin,wang2021pyramid,yuan2021tokens} and local structure modelling~\cite{Swin,dong2021cswin,chen2021visformer}. Further, vision transformers have also been generalized for action recognition and detection in videos~\cite{MViT, Swin, ViViT, MotionFormer, neimark2021video, bertasius2021space}. 
	
	However, a crucial problem with scaling up transformer models is the growth of required GPU memory with depth. This linear growth in memory is prohibitive to the development of very deep models since the batch size needs to be reduced considerably to be able to accommodate storing the intermediate activations on GPU. This problem is exacerbated in video models which process very large input tensors and are often trained with batch size $1$ even for shallower depths. A potential systems-level solution to scale up conventional transformer architectures is model parallelism~\cite{dean2012large} that puts different parts of the model on different GPUs. However in practice, it is quite slow and requires special high bandwith network infrastructure because of huge across device traffic. 
	
	In this work, we use Vision Transformers~\cite{ViT} and Multiscale Vision Transformers~\cite{MViT} as our base models and propose their reversible transformer version that decouple the memory requirement from depth of the model. This facilitates saving GPU memory and allows training with much higher batch size, and consequently, to preserve or even \textit{increase} training throughput of deep non-reversible models. 

\paragraph{Reversible Architectures} are a family of neural network architectures that are based on the NICE~\cite{dinh2014nice, dinh2016density} reversible transformation model which are the precursors of the modern day generative flow based image generation architectures~\cite{ho2019flow++, kingma2018glow}. Based on the NICE invertible transformations, Gomez~\etal~\cite{gomez2017reversible} propose a Reversible ResNet architecture that employs the reversible transformation~\cite{dinh2014nice} for memory-efficient image classification in ResNets~\cite{He2015}. An interesting line of work builds upon the Reversible ResNets ideas proposing better reversible CNN models using ODE characterizations~\cite{chang2018reversible, li2021m, sander2021momentum}, momentum~\cite{li2021m, sander2021momentum}, layer-wise inversion~\cite{hascoet2019layer}, fourier transform based inversion~\cite{finzi2019invertible} and fixed point iteration based inversion~\cite{behrmann2019invertible, song2019mintnet}. Reversible CNNs have been applied to several traditional image tasks such as compression~\cite{liu2021semantics}, reconstruction~\cite{li2020rev}, retrieval~\cite{li2019attention}, and denoising~\cite{huang2021winnet, liu2021invertible} as well as to compressed sensing~\cite{sun2021invertible}, compact resolution~\cite{yang2020image}, image to image translation~\cite{van2019reversible}, remote sensing~\cite{peters2020fully}, medical image segmentation~\cite{pendse2020memory, yamazaki2021invertible} and MRI reconstruction~\cite{putzky2019invert}. Reversible transformation have also been adapted to other networks such as RNNs~\cite{mackay2018reversible}, Unet~\cite{brugger2019partially, etmann2020iunets}, Masked Convolutional Networks~\cite{song2019mintnet} and 1000-layer deep Graph Neural Networks~\cite{li2021training}. Some early attempts have also been made to adapt the reversible transformation to the NLP domain, initiated by Kiatev~\etal\cite{kitaev2020reformer} and built upon in ~\cite{zhao2021multi, zheng2021duplex} for machine translation. 
	
	However, word-level input partitioning contains much richer semantic content than patch level image partitioning and NLP transformers tend to be shallower in depth but wider in channel dimension. For example, Kiatev~\etal\cite{kitaev2020reformer} focus on expanding on the input sequence dimension rather than model depth and with no benchmarking on maximum batch-size, peak GPU memory and training throughput. 
	
	Our experiments show that a na\"ive adaption of reversible vision transformers performs poorly for deeper ($\geq$8 blocks) models. This work is the first to propose Reversible Vision Transformers, adapt it to two state-of-the-art transformer networks, namely, ViT and MViT. Furthermore, this work is the first use of a reversible backbone for object detection and video classification, which tends to be one the most memory starved domains of visual recognition. 
	
	\section{Approach}
	\label{sec:method}
	We first present a brief overview of the reversible transformation (\sref{sec:method:transform}) and its benefits in neural network training (\sref{sec:method:noncaching}). We then present our proposed Reversible Vision Transformer (\sref{sec:method:vit}) its two residual stream structure (\sref{sec:method:vit:res} and associated constraints (\sref{sec:method:noncaching}. This is followed by our proposed Reversible Multiscale Vision Transformer architecture (\sref{sec:method:mvit}) and its sub-blocks (\sref{sec:method:mvit:preserving} and \sref{sec:method:mvit:transition}) that allow end-to-end reversible training. 
	\subsection{Reversible Block Structure}
	The reversible transformer is composed of a stack of reversible blocks that follow the structure of the reversible transformation to allow analytic invertibility of outputs.
	\subsubsection{Reversible Transformation}
	\label{sec:method:transform}
		Consider a transformation $T_1$ that transforms an input tensor $I$ partitioned into two $d$ dimensional tensors, $[I_1; I_2]$ into the output tensor $O$ also similarly partitioned into tensors, $[O_1; O_2]$ with an arbitrary differentiable function $F(\cdot) : \mathbb{R}^d \rightarrow \mathbb{R}^d$ as follows:
	 \[ 
	    \mathbf{I} = \begin{bmatrix}
           I_{1} \\
           I_{2}
         \end{bmatrix} 
         \underset{T_1}{\longrightarrow} 
         \begin{bmatrix}
           O_{1} \\
           O_{2} 
         \end{bmatrix}  
         = 
         \begin{bmatrix}
           I_{1} \\
           I_{2} + F(I_{1})
         \end{bmatrix} =  \mathbf{O}
	\]
	Note that the above transformation $T_1$ allows an inverse transformation $T_1'$ such that  $T_1' \circ T_1$ is an identity transform. Also, consider an analogous transposed transformation $T_2$ using the function $G(\cdot) : \mathbb{R}^d \rightarrow \mathbb{R}^d$ as follows: 
	\[ 
	    \mathbf{I} = \begin{bmatrix}
           I_{1} \\
           I_{2}
         \end{bmatrix} 
         \underset{T_2}{\longrightarrow} 
         \begin{bmatrix}
           O_{1} \\
           O_{2} 
         \end{bmatrix}  
         = 
         \begin{bmatrix}
           I_{1} + G(I_{2}) \\
           I_{2} 
         \end{bmatrix} =  \mathbf{O}
	\]
	Similar to  $T_1$,  $T_2$ also allows an inverse transform  $T_2'$. Now consider the composition $T = T_2 \circ T_1$ that transforms both the partitions of the input vector $\mathbf{I}$ and is obtained as, 
	\begin{equation}
	    \mathbf{I} = \begin{bmatrix}
           I_{1} \\
           I_{2}
         \end{bmatrix} 
         \underset{T}{\longrightarrow} 
         \begin{bmatrix}
           O_{1} \\
           O_{2} 
         \end{bmatrix}  
         = 
         \begin{bmatrix}
           I_{1} + G(I_{2} + F(I_1)) \\
           I_{2} + F(I_1)
         \end{bmatrix} =  \mathbf{O}
    \end{equation}
    \label{eq:transform}
	Naturally, $T$ affords the inverse transform $T' = T_1' \circ T_2'$ that follows $T'(T(\mathbf{I})) = \mathbf{I}$. Note that the inverse transform $T'$ queries the functions $F$ and $G$ exactly once and hence has the same computational cost as the forward transform $T$.
	\subsubsection{Vanilla networks require caching activations}
	\label{sec:method:caching}

Consider the back-propagation mechanism. Given a computation graph node, $\mathcal{M}$, its children nodes $\{\mathcal{N}_j\}$, and the gradients of the children node with respect to final loss $\cbr{\frac{d \mathcal{L}}{d \mathcal{N}_j}}$, the back-propagation algorithm uses the chain rule to calculate the gradient with respect to $\mathcal{M}$ as, 
\[
    \frac{d \mathcal{L}}{d \mathcal{M}} = \sum_{\mathcal{N}_j}\left(\frac{\partial f_j}{\partial \mathcal{M}} \right)^T\frac{d \mathcal{L}}{d \mathcal{N}_j}
\]
where $f_j$ denotes the function computing node $\mathcal{N}_j$ from its parents, $\mathcal{M}$ being one of them. The jacobian $\frac{\partial f_j}{\partial \mathcal{M}}$, requires calculating the partial gradient of the $f_j$ output with respect to the current node $\mathcal{M}$. 
    
Now consider the simplest possible neural network layer $f(X) = W^TX$, where $X$ is an intermediate activation inside the network. Applying the above described backpropagation algorithm to compute the derivative with respect to parent nodes, and using the output $Y$ as the sole child node, $\mathcal{N}_j$, we get, 
\begin{align*}
\frac{d \mathcal{L}}{d W} &= \left(\frac{d \mathcal{L}}{d Y}\right)X^T & 
\frac{d \mathcal{L}}{d X} &= W\frac{d \mathcal{L}}{d Y}
\end{align*}

Thus, because of the function jacobian, the backpropagation algorithm requires intermediate activations during the forward pass to be available in the backward pass to compute the gradients with respect to the weights. 

Typically, this is achieved by caching the intermediate activations on GPU memory for use in the backward pass. This allows fast gradient computation at the cost of extra memory. Further, the sequential nature of the network requires the activations for all the layers to be cached in before the loss gradients are calculated and the cached memory is freed. This dependence significantly affects the peak memory usage which thus becomes linearly dependent on the network depth $D$. 

% %%%%%%%%%%%%%%%%%%%%%%%%%%%%%%%%%%%%%%%%%%%%%%%%%%%%%%%%%%%%%%%%%%%%%%%%%%%%%%%
\begin{figure*}[t]
    % left bottom top right
  \vspace{-15pt}
    	\centering
	\subfloat[
	Rev-\textbf{ViT} Block\label{fig:rev_vit}
	]{%no unwanted space
		\includegraphics[width = 0.255\textwidth,  clip]{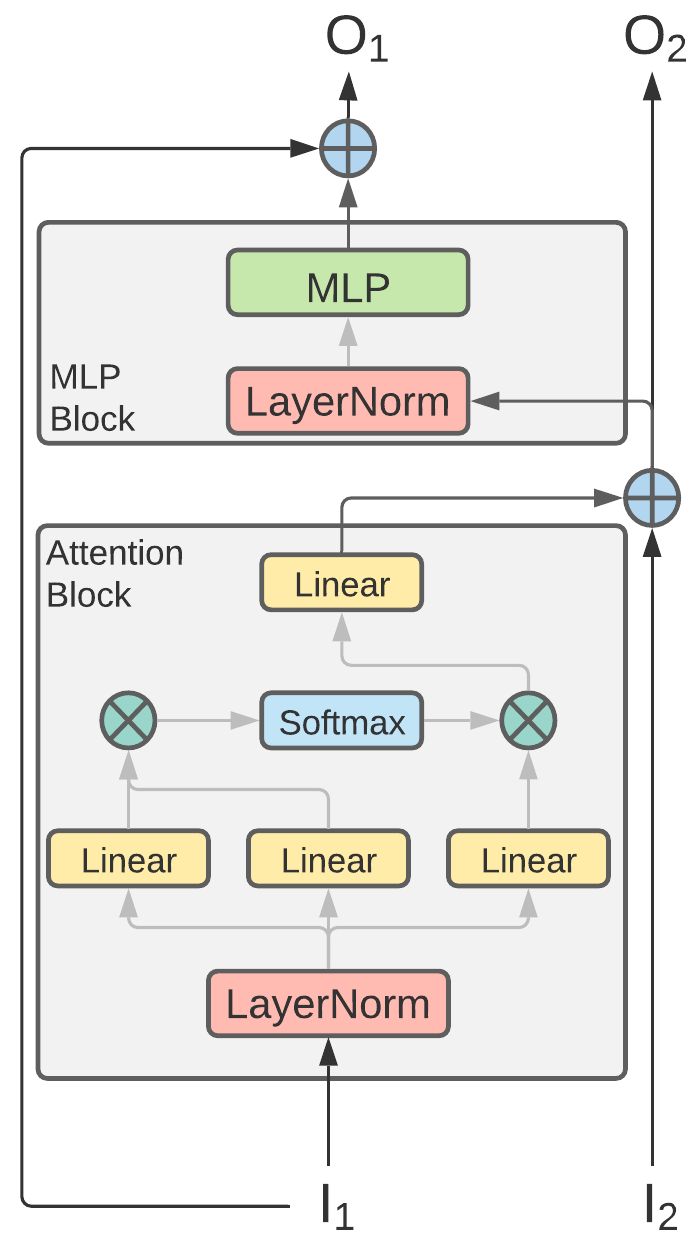}
	}\hfill
	\subfloat[
	Stage-Transition Rev-\textbf{MViT} Block\label{fig:rev_mvit_1}
	]{%no unwanted space
		\includegraphics[width = 0.26\textwidth,  clip]{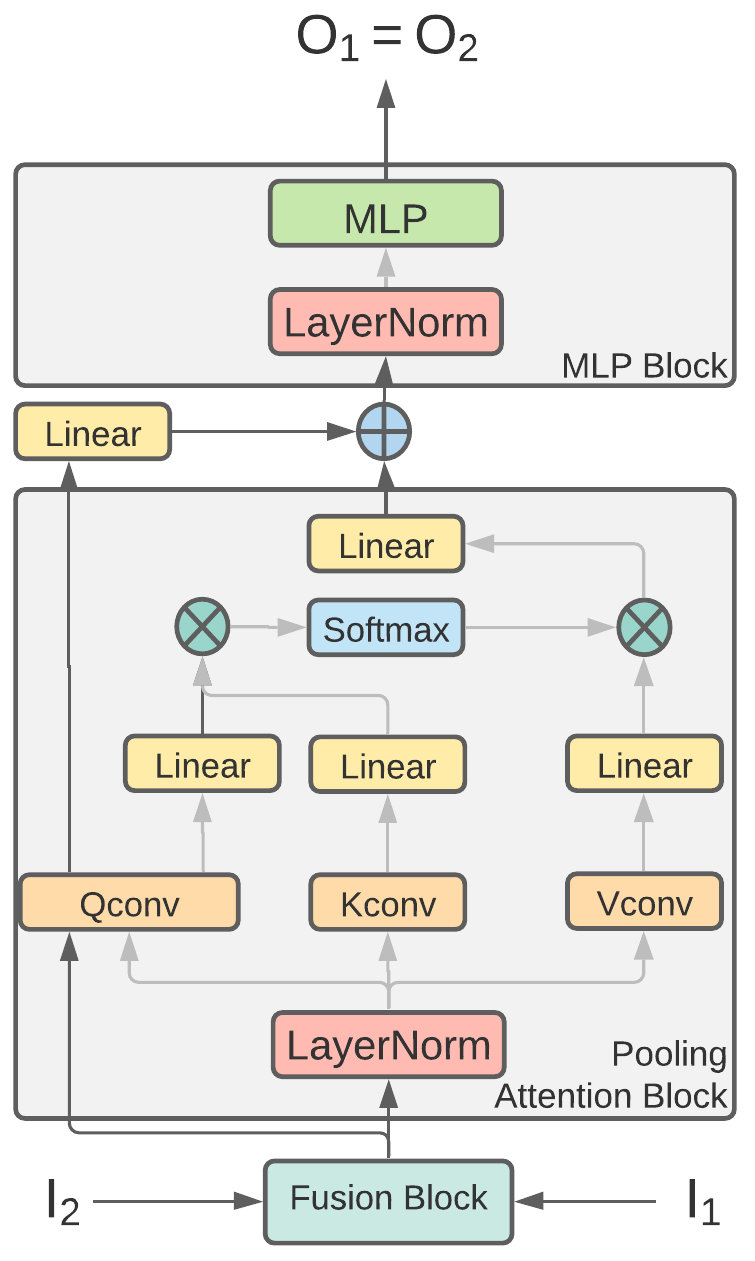}
	}\hfill
	\subfloat[
	Stage-Preserving Rev-\textbf{MViT} Block\label{fig:rev_mvit_2}
	]{%no unwanted space
		\includegraphics[width = 0.33\textwidth, clip]{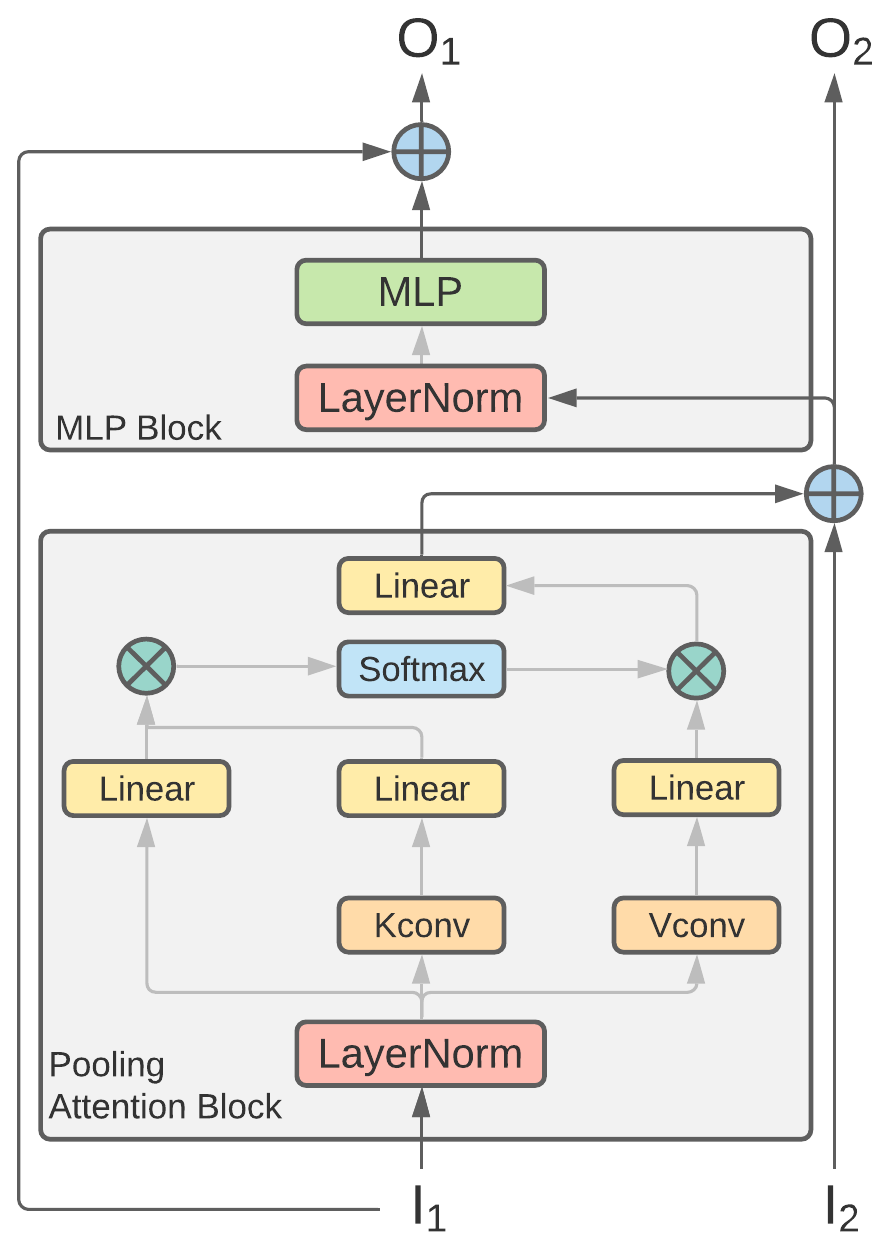}
	}
	\caption{\textbf{Reversible ViT} is a two-residual-stream architecture composed of a stack of Reversible ViT blocks \protect\subref{fig:rev_vit} that transforms the inputs $I_1$ and $I_2$ with the ViT design~\cite{ViT}, but in our reversible fashion. \textbf{Reversible MViT} is a two-residual-stream architecture as well, made up of a stack of two type of blocks -- 
  \protect\subref{fig:rev_mvit_1}  The stage-transition blocks that act as coupling between the residual streams as well as perform channel upsampling and resolution downsampling and \protect\subref{fig:rev_mvit_2} the stage-preserving blocks that form the majority of the computational graph and propagate information preserving input feature dimension.}
    \label{fig:rev_all}
    \vspace{-10pt}
\end{figure*}
% %%%%%%%%%%%%%%%%%%%%%%%%%%%%%%%%%%%%%%%%%%%%%%%%%%%%%%%%%%%%%%%%%%%%%%%%%%%%%%%

\subsubsection{Learning without caching activations}
\label{sec:method:noncaching}
As noted in \sref{sec:method:transform}, an input transformed with the reversible transformation $T$ allows recalculating the input from the output of the transformation. Hence, a network composed of such reversible transformations \textit{does not need to store intermediate activations} since they can be recomputed easily in the backward pass from the output. However the reversible transformation $T$ places an important constraint on the property of the learnt function.   

\noindent \textbf{Equidimensional Constraint}. As mentioned in \sref{sec:method:transform}, the functions $F$ and $G$ need to be equidimensional in input and output spaces. Hence, the feature dimensions need to remain constant under $T$. While this constraint is an obstruction for other vision architectures such as ResNets \cite{He2015} that require a change of feature dimensions, it is easily satisfied in the Vision Transformer architecture \cite{ViT} which maintains a constant feature dimension throughout the layers.  
\subsection{Reversible Vision Transformers}
\label{sec:method:vit}

\subsubsection{Adapting ViT to Two-Residual-Streams}
\label{sec:method:vit:res}
Fig.~\ref{fig:rev_vit} shows the reversible transformation $T$ adapted to the Vision Transformer architecture \cite{ViT}. The input consists of two partitioned tensors $I_1$ and $I_2$ that are transformed as per the equation \ref{eq:transform} maintaining reversibility. This leads to a \textit{two-residual-stream} architecture where each of the inputs $I_1$ and $I_2$ maintain their own residual streams while mixing information with each other using functions $F$ and $G$. Following ViT \cite{ViT}, we use the Multi-head attention and the MLP sublocks as functions $F$ and $G$ respectively. 

\subsubsection{Boundary Conditions}
\vspace{-5pt}
\label{sec:method:vit:boundary}
As the ViT architecture only uses a single residual stream, the architecture needs to be modified to support the two-residual-stream design (\sref{sec:method:vit:res}). We propose the following:

\noindent\textbf{1. Initiation.} We keep the stem intact and send the patchification output activations to $I_1$ and $I_2$. Note that this design choice is different from \cite{rev_resnets} which proposes to split in halves along the channel dimensions. 

\noindent\textbf{2. Termination.} The two residual paths need to be fused before the final classifier head to preserve information. We propose to layer-normalize the inputs first, followed by concatenation, to reduce the fusion computational overhead.   

\subsubsection{Reconfiguring Residual Connections}
\label{sec:method:vit:drift}
Residual connections play a key role for signal propagation in deep networks~\cite{He2015}. The reversible transform $T$ itself also depends crucially on the residual connections between the two streams to maintain reversibility. Interestingly, we observe a key relationship between the residual connections and signal propagation in Reversible Vision Transformer. 

Note that while it is common practice for neural network blocks to be wrapped around a residual connection for better gradient flow~\cite{He2015}, there is no such connection for either the $I_1$ or $I_2$ inputs. Specifically, internal residual connections around the MLP and attention sub-blocks for both the $I_1$ and $I_2$ streams are absent. Instead, the residual connections for each residual stream flows through the other stream, operating through the inherent skip connection present in the reversible transformation $T$ (\sref{sec:method:transform}). We find these \textit{internal} skip connections \textit{detrimental} to training convergence for deeper models while bringing no additional gain for shallower models and choose to omit them entirely for reversible vision transformer blocks. 

\subsection{Reversible Multiscale Vision Transformers}
\label{sec:method:mvit}
The recently proposed MViT architecture develops a feature hierarchy inside the model by \textit{downsampling} the visual \textit{resolution} and \textit{upsampling} the \textit{channel} dimension. It obtains state-of-the-art results on both image and video classification benchmarks. To showcase the flexibility of the reversible design, we adapt the MViT model to Reversible Multiscale Vision Transformers. We propose to compose the Reversible MViT architecture in the same structure as the MViT model but using two different layers --  the \textit{Stage Transition} and the \textit{Stage-Preserving} blocks.  
\subsubsection{Stage-Transition Block}
\label{sec:method:mvit:transition}
Figure \ref{fig:rev_mvit_1} depicts the architecture of the proposed stage-transition block. The stage-transition block closely follows the design of the resolution upsampling blocks in MViT \cite{MViT} with the following crucial modifications:
\paragraph{Lateral Connections.} The residual streams $I_1$ and $I_2$ are fused with lateral connections at the start of the stage-transition block. This allows efficient computation of the resolution downsampling and feature upsampling without repeat computation in each stream separately.  
\paragraph{Feature Upsampling.} MViT performs feature upsampling in the last MLP block before the resolution upsampling block. We propose to move the channel upsampling stage inside the pooling attention sub-block of the stage-transition block. Specifically, we propose to upsample the Query, Key and Value vectors in the linear layer following the pooling channel-wise convolutional layers (Figure \ref{fig:rev_mvit_1} and \ref{fig:rev_mvit_2}). This was the dual benefit of (A) allowing all feature dimension changes to take place in sync inside the same block and allowing other blocks to keep feature dimensions intact, a virtue of reversible architectures (\sref{sec:method:noncaching}) and (B) saving additional computation from being used in the prior MLP and pooling layers. 
We follow the same boundary conditions at the stage-transition blocks as in the reversible vision transformer architecture (\sref{sec:method:vit:boundary}). 
\subsubsection{Stage-Preserving Block}
\label{sec:method:mvit:preserving}
Figure \ref{fig:rev_mvit_2} shows the reversible transformation $T$ (\sref{sec:method:transform}) adapted to the Multiscale Vision Transformer architecture~\cite{MViT}.  The design closely resembles that of the reversible vision transformer block (Figure \ref{fig:rev_vit}) with the addition of multi-head pooling attention~\cite{MViT}. Note that even though the attention uses pooling on key and value tensors, thereby changing the sequence length, the output dimensions are still preserved. Hence, the stage-preserving block still follows the equidimensional constraint (\sref{sec:method:noncaching}) and hence can be made fully reversible and learnt without caching activations. 

Since each stage-transition block changes the spatiotemporal resolution, they occur only a limited number of times in the entire MViT network. In other words, the majority of the computation as well as memory usage is performed within the stage-preserving blocks and is fully reversible. We follow the same residual connection circuit (\sref{sec:method:vit:drift}) as in Reversible Vision Transformer blocks for both the stage-transition and the stage-preserving blocks. 
\section{Results}

\paragraph{Datasets.}  We benchmark both the Reversible Vision Transformer and the Reversible Multiscale Vision Transformer architectures extensively across image classification (ImageNet \cite{deng2009imagenet}), video classification (Kinetics 400~\cite{Kay2017} \& Kinetics 600~\cite{Carreira2017}) and object detection (MS-COCO \cite{Lin2014}). Across all the benchmarks, we observe significant memory savings by using the reversible architecture with negligible to no accuracy change. All presented results and ablations are trained from random initialization, except for COCO where we initialize from ImageNet weights.

\subsection{Image Classification}
\label{sec:results:IN}
\paragraph{Settings.} We benchmark our proposed models on image classification on the ImageNet-1K dataset~\cite{deng2009imagenet} with \app1.28M images among 1000 classes. We follow training recipes~\cite{fan2020pyslowfast} for both ViT~\cite{ViT} and MViT~\cite{MViT} models with certain crucial adaptions (\sref{tab:ablation:fusion}). All models are trained from random initialization without EMA for 300 epochs except for ViT-L and Rev-ViT-L which follow a 200 epoch training recipe. Training details are in Supplementary. 
\paragraph{Results.} Table \ref{tab:sota:in1k} shows the results for Reversible Vision and Reversible Multiscale Vision Transformers across different models and FLOP regimes. We benchmark all the models on a single 16 GB V100 GPU under $224\times224$ image size and otherwise identical conditions. The maximum batch size is obtained as the highest number of images in a batch than can train without running out of GPU memory. The memory per image is measured as the peak GPU memory each image occupies during training. 

We note that Reversible Vision Transformers match the FLOP and parameter specifications of their non-reversible counterparts exactly owing to the parsimonious design of the reversible vision transformer block (\sref{sec:method:vit}). 
The Reversible Multiscale Vision Transformer has slightly higher FLOPs due to the stage-transition (\sref{sec:method:mvit:transition}) stages while still being very GPU memory efficient owing to the stage-preserving (\sref{sec:method:mvit:preserving}) stages.

%##################################################################################################
\begin{table}[t!]
        \centering
        \tablestyle{1.8pt}{1.05}
        	\resizebox{1.04\linewidth}{!}{
        \begin{tabular}{l|c|l|l|r|r}
            \multicolumn{1}{c|}{model}  &   Acc  & \specialcell{\textbf{Memory} \\ (MB/img)}  & \specialcell{\textbf{Maxiumum} \\ \textbf{Batch Size}} & GFLOPs & Param (M) \\
            \shline
        
            {ResNet-101 \cite{He2016a}}                                      &  76.4  & 118.7 & 112 &  7.6   & 45 \\
            {ResNet-152 \cite{He2016a}}                                       & 77.0 & 165.2 & 79 & 11.3   & 60 \\
            {RegNetY-4GF \cite{ilija_2020} }                                 &  80.0  & 101.1  & 136 & 4.0 & 21   \\ 
            {RegNetY-12GF \cite{ilija_2020} }                                &  80.3 & 175.2 & 75 & 12.1   &  51.8 \\
            {RegNetY-32GF \cite{ilija_2020} }                                 &  80.9  & 250.2  & 46 & 32.3 & 32.3   \\ 
            {Swin-T~\cite{liu2021swin}}                                       &  81.3 & - & - &  4.5   & 29  \\
            \shline
            ViT-S \cite{deit}                                &  79.9 & 66.5 & 207 & 4.6 & 22 \\
            \rowcolor{baselinecolor}
            Rev-ViT-S                                                        & 79.9 & \save{\textbf{8.8}}{\textbf{7.5}\x} & \increase{\textbf{1232}}{\textbf{5.9}\x} & 4.6 & 22\\
            ViT-B \cite{deit} &      81.8 & 129.7 & 95 & 17.6 & 87 \\
            \rowcolor{baselinecolor}
            Rev-ViT-B  &  81.8 & \save{\textbf{17.0}}{\textbf{7.6}\x} & \increase{\textbf{602}}{\textbf{6.3}\x}  & 17.6 & 87 \\
            \shline
            {RegNetY-8GF \cite{ilija_2020} }                                 &  81.7 & 147.2 & 91 &  8.0   & 39 \\
            CSWin-T~\cite{dong2021cswin}                                     &  82.7 & - & - &  4.3   & 23 \\
            Swin-S~\cite{liu2021swin}                                        &  83.0 & - & - &  8.7   & 50 \\
            \shline
            ViT-L &      81.5 & 349.3 & 26 & 61.6  & 305 \\
            \rowcolor{baselinecolor}
            Rev-ViT-L  &  81.4 & \save{\textbf{22.6}}{\textbf{15.5}\x} & \increase{\textbf{341}}{\textbf{13.1}\x}  & 61.6 & 305  \\
            MViT-B-16~\cite{MViT} &     82.8 & 153.6 & 89 & 7.8 & 37 \\
            \rowcolor{baselinecolor}
            Rev-MViT-B-16  &           82.5 & \save{\textbf{66.8}}{\textbf{2.3}\x} & \increase{\textbf{157}}{\textbf{1.8}\x} & 8.7 & 39  \\
            
        \end{tabular}
        }
        
        \caption{\textbf{Comparison to prior work on ImageNet-1K classification}. All memory and maximum batch size are on 224$\times$224 input resolution on a 16G V100 GPU. \textbf{Rev-ViT} and \textbf{Rev-MViT} match performance across different FLOP regimes at a fraction of the per-input GPU memory cost.
        }
        \label{tab:sota:in1k}
    \end{table}
%##################################################################################################

\paragraph{Increasing memory savings with depth.}  In Table \ref{tab:sota:in1k}, we observe that our Rev-ViT matches the performance of vanilla ViT to very close fidelity across all model variants (Small, Base and Large) and FLOP regimes. Since the memory used per image is linearly dependent on the depth of the model for vanilla networks (\sref{sec:method:caching}), the memory gains of the reversible model increases as the network scales in depth. Notably, while the Reversible ViT-S already enjoys an impressive memory saving of about \textbf{86.8}\% (equivalent to a $\textbf{7.6}$\x\ reduction) with respect to the vanilla ViT-S model, the gain increases further to $\textbf{15.5}$\x\ or, about $\textbf{93}$\% memory savings for the Reversible ViT-L model.

Equivalently, the saved memory can be used to increase the training batch size where we observe a similar trend as well. While reversible ViT-S model achieves a $\textbf{6.1}$\x\ increase in batch size on the ViT-S model, the effect is more for ViT-L model where the maximum batch size increases by $\textbf{14.3}$\x\, jumping from a small $24$ image per batch to $344$ images. This is a very favorable trend, since it is indeed the deeper models that hit the \textit{GPU memory wall}~\cite{gholami2021}.

Further, hierarchical vision transformers such as MViT also enjoy a memory saving of about $\textbf{52.1}$\% without suffering any significant drop in performance. The memory savings in Rev-MViT are smaller compared to the ViT variants because of the stage-transition blocks in hierarchical models (\sref{sec:method:mvit:transition}) that require storing the input activations due to the non-reversible nature of pooling attention stemming from the feature dimension change~\cite{MViT}. 

%%%%%% new table %%%
\begin{table}[t!]
	\hspace*{-7pt}
	\centering
	\tablestyle{2.0pt}{1.04}
	\begin{tabular}{l|c|x{15}x{15}|r|r}
		\multicolumn{1}{c|}{\multirow{2}{*}{model}}  &\multicolumn{1}{c|}{\multirow{2}{*}{top-1}} & {\textbf{Mem} (GB)}  & {\textbf{Max BS}}  &\multirow{2}{*}{\specialcell{\scriptsize GFLOPs\x\\~views}} & \multirow{2}{*}{Param} \\
		\shline
		\hline		
		Two-Stream I3D \cite{Carreira2017}&    71.6 & - & - & 216~\x~NA & 25.0 \\
		R(2+1)D \cite{Tran2018}&   72.0 & - & - &152\x115  & 63.6 \\
		Two-Stream R(2+1)D \cite{Tran2018} &  73.9 & - & - & 304~\x~115 & 127.2  \\
		Oct-I3D + NL \cite{chen2019drop}&   75.7  & - & - &28.9\x3\x10 &  33.6  \\
		ip-CSN-152  \cite{Tran2019}&  77.8 & - & - & 109\x3\x10 & 32.8  \\
		{SlowFast} 4\x 16, R50 \cite{Feichtenhofer2019}& 75.6  & - & - &36.1~\x~30 & 34.4 \\
		{SlowFast} 8\x 8, R101 \cite{Feichtenhofer2019}& 77.9  & - & - &  106~\x~30 & 53.7  \\
		{SlowFast} {\scriptsize 8\x 8 +NL}  \cite{Feichtenhofer2019}& {78.7} & - & - &  116\x3\x10 & 59.9  \\ 
		\shline
		{ViT-B-VTN-{IN-1K} \cite{neimark2021video}} &    \demph{75.6} & - & - &  {4218\x1\x1} &  {114.0} \\
		{ViT-B-VTN-{IN-21K} \cite{neimark2021video}}  &    \demph{78.6} & - & - & {4218\x1\x1} &  {114.0} \\
		\shline
		MViT-B-16 , 16\x4 & 78.4 & 1.27 & 10 & 70.5\x1\x5& 36.6 \\ 
		\rowcolor{baselinecolor}
		\textbf{Rev-MViT}-B-16, 16\x4 & 78.5 & \textbf{0.64} & \textbf{20} & 64\x1\x5& 34.9 \\ 
	\end{tabular}
	\caption{\textbf{Comparison to prior work on Kinetics-400 video classification}. Single view inference cost is reported along with used number of views (FLOPs\x view$_\text{space}$\x view$_\text{time}$). Memory (Mem) reported in Gigabytes per input clip. Maximum Batch Size (Max BS) measured as the maximum possible single GPU batch size. All measurements are performed on a single 16G V100 GPU.}
	\label{tab:sota:k400}
\end{table}
%##################################################################################################

%##################################################################################################
\begin{table}[t!]
	\centering
	\small
	\tablestyle{2.0pt}{1.04}
	\begin{tabular}{l|c|x{15}x{15}|r|r}
		\multicolumn{1}{c|}{\multirow{2}{*}{model}}  &\multicolumn{1}{c|}{\multirow{2}{*}{top-1}} & {\textbf{Mem} (GB)}  & {\textbf{Max BS}}  &\multirow{2}{*}{\specialcell{\scriptsize GFLOPs\x\\~views}} & \multirow{2}{*}{Param} \\
		\shline
		{SlowFast} {\scriptsize{16\x 8 +NL}} \cite{Feichtenhofer2019} & {81.8}  & - & - &  234\x3\x10 & 59.9  \\
		{X3D-XL} & {81.9}  & - & - & 48.4\x3\x10 & 11.0  \\
		ViT-B-TimeSformer-IN-21K~\cite{bertasius2021space} & 82.4 & - & - &  1703\x3\x1 &  121.4  \\
		ViT-L-ViViT-IN-21K~\cite{ViViT} & 83.0 & - & - &  3992\x3\x4 & 310.8 \\ 

		\hline
		MViT-B-16, 16\x4 & 81.3 & - & - & 70.3\x1\x5   & 36.6 \\ % 
		MViT-B-16, 32\x3 & 83.4 & - & - & 170\x1\x5  & 36.8 \\ %   
		\hline 
		MViT-B-24, 32\x3 & 83.8 & 4.40 & 2 &  236\x1\x5 & 52.9  \\ 
		\rowcolor{baselinecolor}
		\textbf{Rev-MViT}-B-24, 32\x3 & 83.7 & \textbf{1.64} &\textbf{7} &   223\x1\x5 & 51.8 \\ 
	\end{tabular}
	\caption{\textbf{Comparison to prior work on Kinetics-600 video classification}. Results under same settings as Kinetics-400 in Table \ref{tab:sota:k400}.
	}
	\label{tab:sota:k600}
\end{table}

\subsection{Video Classification}
\paragraph{Settings.} We also benchmark Rev-MViT-B models on action classification on Kinetics-400~\cite{Kay2017} and Kinetics-600~\cite{Carreira2017} datasets. All the models are trained {from scratch} with training recipes adapted from \cite{MViT}. 

\paragraph{Results.} Table \ref{tab:sota:k400} and \ref{tab:sota:k600} present the results on action recognition task on the Kinetics-400~\cite{Kay2017} and Kinetics-600~\cite{Carreira2017} respectively. For action recognition, we benchmark our adapted Reversible  MViT model and report top-1 and top-5 performance for both datasets. Similar to the image classification benchmark, we observe that the Reversible MViT models closely match the accuracy for their non reversible counterparts at a fraction of the memory cost.

\paragraph{Increasing video model batch sizes.} 
We note that our adapted Reversible MViT forms a very competitive video recognition model. Specifically,  for both Kinetics-400 (Table \ref{tab:sota:k400}) and Kinetics-600 (Table \ref{tab:sota:k600}) the reversible models match the overall accuracy very closely at only $\textbf{51.5}\%$ and $\textbf{37.2}\%$ of the memory cost respectively.     

This allows a batch size increase of $\textbf{2}$\x\ on the 16 layer, 70.5 GFLOPs Kinetics-400 MViT-B-16 model and of $\textbf{3.5}$\x\ on the 24 layer, 236 GFLOPs Kinetics-600 MViT-B-24 model, a very beneficial result for large video models which are often severely memory limited and trained with very small batch sizes (Table \ref{tab:sota:k600}). Moreover, due to the more efficient design of stage-transition blocks in Rev-MViT (\sref{sec:method:mvit:transition}), \ie bringing the dimension upsampling operation inside the pooling attention instead of being performed in the prior MLP stage~\cite{MViT}, the Rev-MViT are also slightly more parameter and FLOP efficient on both Kinetics.

\subsection{Object Detection}
We benchmark the proposed Rev-ViT-B and Rev-MViT-B models on object detection on MS-COCO~\cite{Lin2014} as well. All the models are trained on 118K training images and evaluated on the 5K validation images. We take the ViT-B and MViT-B backbones pre-trained on IN and use the standard Mask R-CNN~\cite{He2017} as the detection framework. All models are trained with a standard 3\x\ schedule (36 epochs). For MViT, we integrate the multi-scale backbone with the feature pyramid network~\cite{lin2017feature}. Referring to Table \ref{tab:COCO} we observe that, the Rev-MViT-B model closely matches the AP performance on MViT-B at only 54.8\% of the memory cost. 

\begin{table}[t!]
    \vspace{-10pt}
\tablestyle{1.0pt}{1.05}
         \hspace{-10pt}
         \small
	    \tablestyle{2.0pt}{1.08}
         \begin{tabular}{l|cc|crr}
             Model & {AP$^\text{box}$} & {AP$^\text{mask}$} & \textbf{Memory}(GB) & GFLOPs  & Param (M)  \\
              \shline
              Res50~\cite{He2016}               & 41.0 & 37.1   & - & 260 & 44 \\
              Res101~\cite{He2016}                & 42.8 & 38.5   & - & 336 & 63\\
              X101-64~\cite{Xie2017}           & 44.4 & 39.7 & - & 493 & 101\\
              PVT-L\cite{wang2021pyramid}      & 44.5 &  40.7 & - & 364 & 81 \\
              \shline
        MViT-B                                &  48.2   & 43.9 & 18.9 & 668 & 57 \\
        \rowcolor{baselinecolor}
        \textbf{Rev-MViT}-B                            &  48.0  & 43.5	 & 10.9 & 683 &	58 \\
          \end{tabular}
    \vspace{-6pt}
          
          \caption{\textbf{Comparison on MS-COCO object detection.} Rev-MViT achieves competitive performance to MViT across all metrics at $\textbf{1.7}$\x\ lower memory footprint.}
          \label{tab:COCO}
    \vspace{-10pt}
\end{table}
%%%%%%%%%%%%%%%%%%%%%%%%%%%%%%%%%%%%%%%%%%%%%%%%%%%%%%%%%%%%%%%%%%%%%%%%%%%%%%%

\subsection{Ablations}
\label{sec:ablations}

%%%%%%%%%%%%%%%%%%%%%%%%%%%%%%%%%%%%%%%%%%%%%%%%%%%%%%%%%%%%%%%%%%%%%%%%%%%%%%%
\begin{figure*}[t!]
\centering
\subfloat[{Activation caching and internal residuals.}\label{fig:ablation:residual}]{%no unwanted space
	\includegraphics[width = 0.33\textwidth]{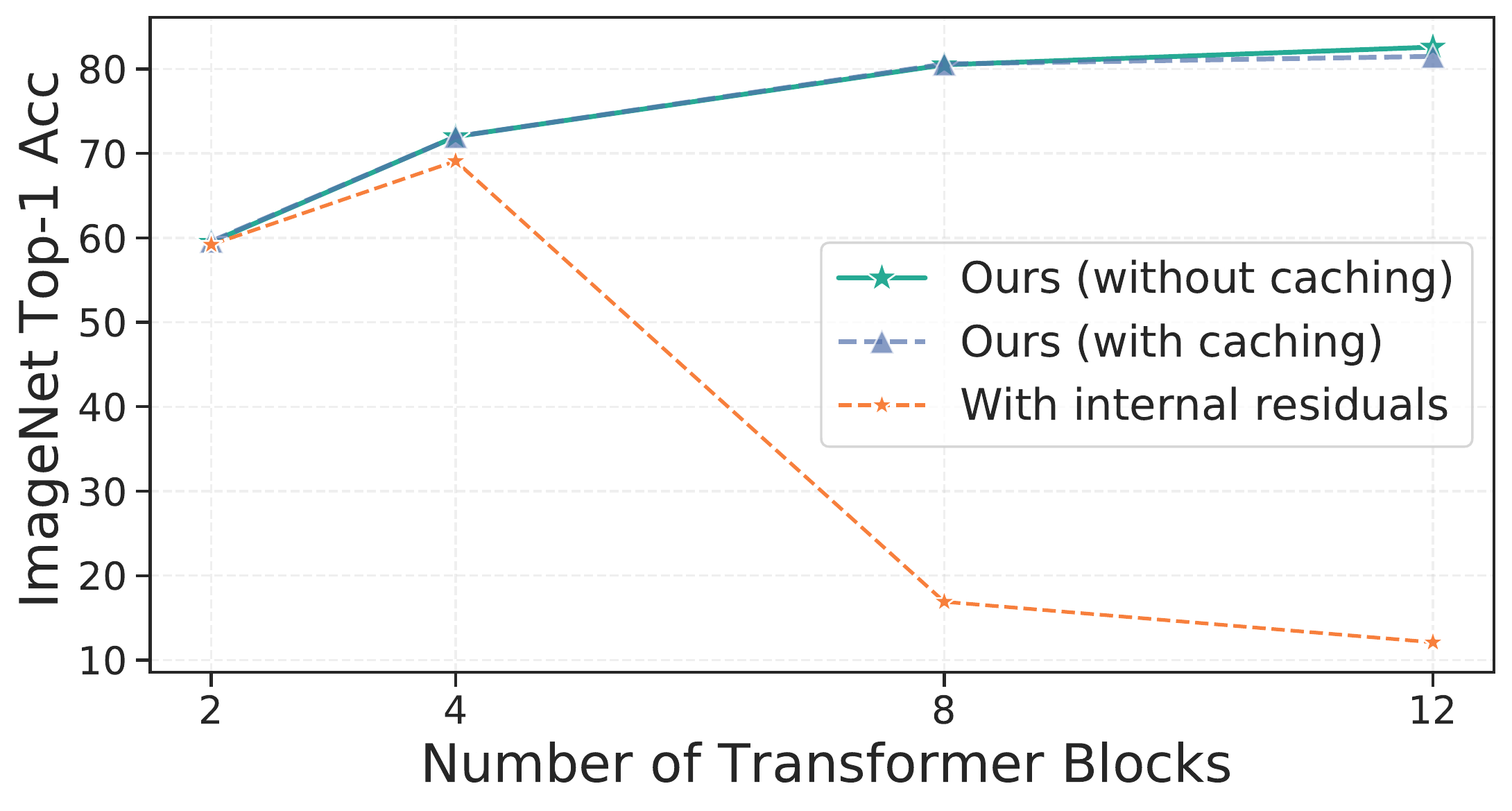}
}
\subfloat[{Training throughput vs. Model Depth}\label{fig:ablation:throughput}]{%no unwanted space
	\includegraphics[width = 0.33\textwidth, trim = 4.0cm 2.6cm 4.0cm 0cm, clip]{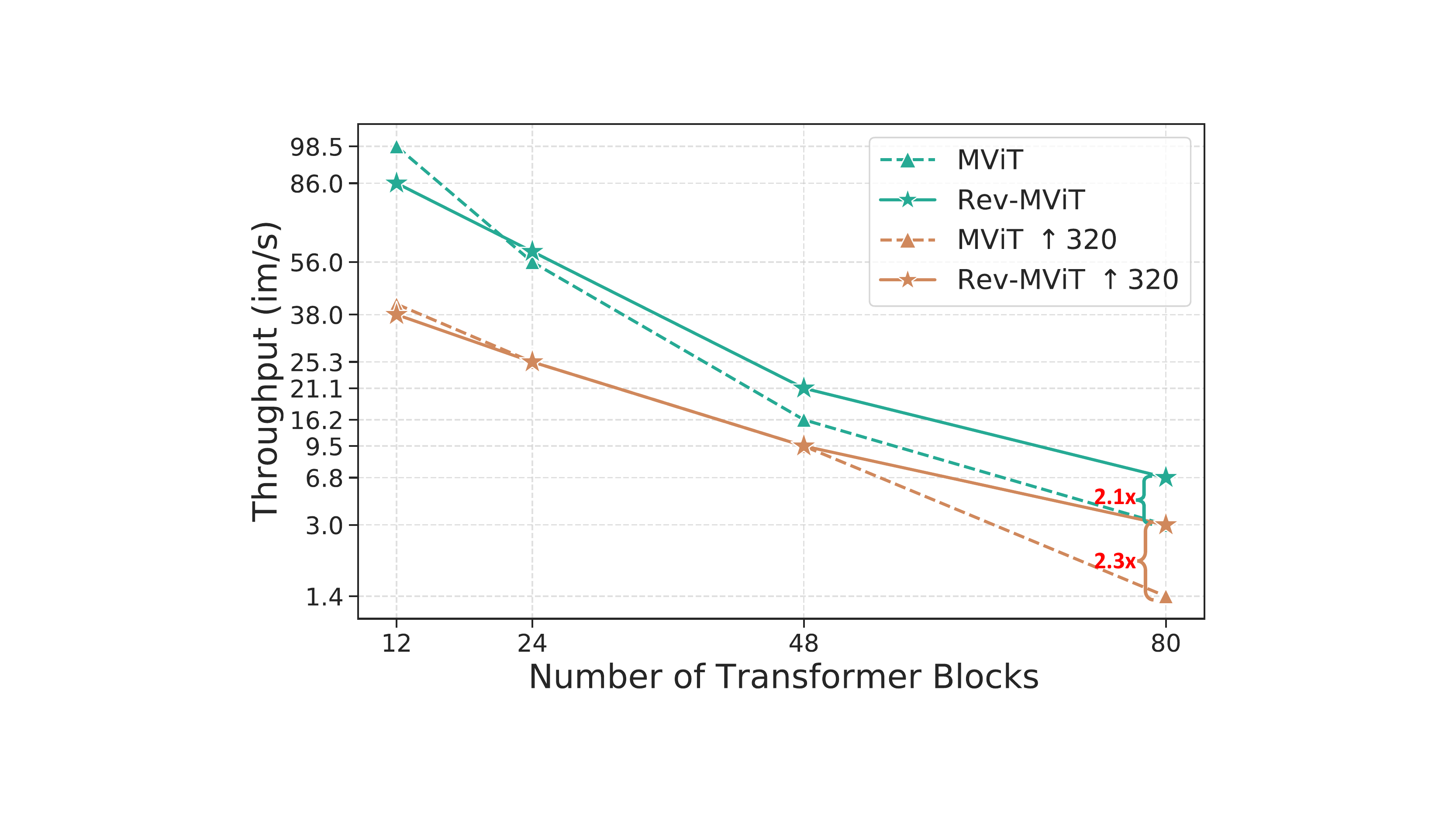}
}
\subfloat[{Reversible training and maximum batch size.}\label{fig:ablation:bscdepth}]{%no unwanted space
	\includegraphics[width = 0.33\textwidth]{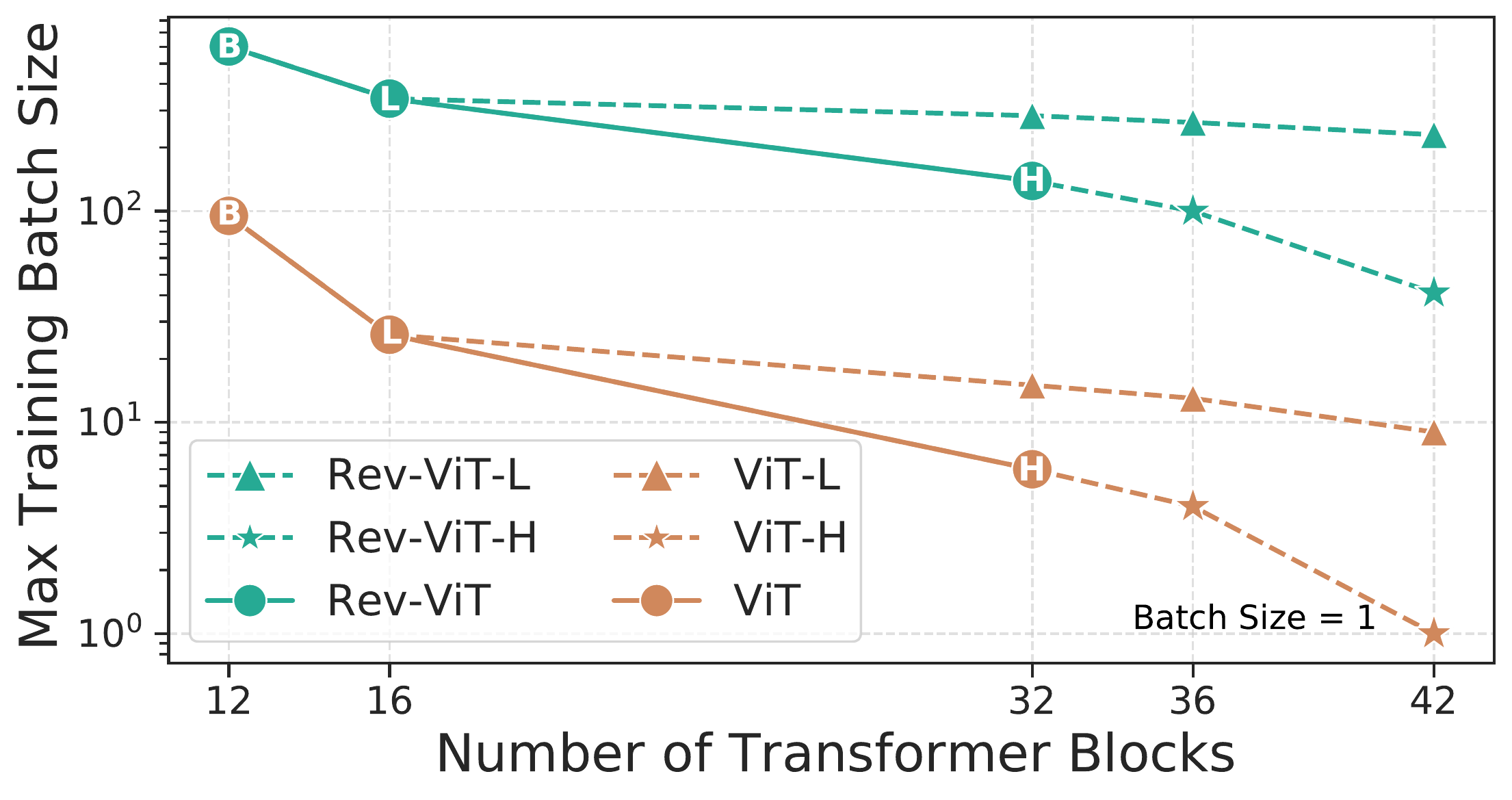}
}
\caption{\textbf{Ablation Experiments.} \protect\subref{fig:ablation:residual}:~We observe that (1) Learning without activation caching does not hurt reversible accuracy for Rev-ViT-B of varying depths and (2) Internal residual connections train well for shallow models but the training \textit{diverges} for deeper models. \protect\subref{fig:ablation:throughput}:~Rev-MViT has higher \textit{throughput} for higher input resolution and deeper MViT models increasing up to \textbf{2.3}\x~at 224 resolution for 80 layers. \protect\subref{fig:ablation:bscdepth}:~We benchmark the maximum batch size for Rev-ViT Base (B), Large (L) and Huge (H) and their non-reversible counterparts.}
\vspace{-10pt}
\end{figure*}
%%%%%%%%%%%%%%%%%%%%%%%%%%%%%%%%%%%%%%%%%%%%%%%%%%%%%%%%%%%%%%%%%%%%%%%%%%%%%%%

\paragraph{Stronger Inherent Regularization.} Across different models and datasets, we find that at the same FLOP and parameter specifications, the reversible models tend to have stronger inherent regularization than their non-reversible counterparts. Hence, training recipes for reversible vision transformers have lighter repeated augmentations, smaller augmentation magnitudes and consequently, higher weight decay. Table ~\ref{tab:ablation:regularization} shows the effects of these recipe changes on Rev-ViT-B. We also observe similar effects on other Rev-ViT and Rev-MViT models where a modified training recipe with lighter augmentations and higher weight decay play a crucial role in matching performance. 
%%%%%%%%%%%%%%%%%%%%%%%%%%%%%%%%%%%%%%%%%%%%%%%%%%%%%%%%%%%%%%%%%%%%%%%%%%%%%%%
\begin{table}[t!]
\vspace{-10pt}
    \centering
    \tablestyle{2.2pt}{1.05}
    \begin{tabular}{l|c|c}
        Training Improvement &  Train Acc & Top-1 ImageNet Acc     \\
        \shline
        Na\"ive Rev-ViT-B              & 15.3 & 12.1  \\
        + Re-configuring residual streams & 82.1 &  77.2  \\ 
        + Repeated Augmentation           & 84.9 & 80.6 \\ 
        + Lighter Augmentation magnitude  & 93.2 & 81.0  \\
        + Stronger Stochastic Depth & 92.0 & 81.4 \\ 
        + Higher weight decay    & 91.0 &  81.8\\
        \hline
        \textbf{Rev-ViT}-B & 91.0 & 81.8 
    \end{tabular}
    \vspace{-6pt}
    \caption{\textbf{Rev-ViT-B Training Recipe.} We observe that reversible transformers tend to have a stronger inherent regularization and require a lighter augmented training recipe for peak performance.}
    \label{tab:ablation:regularization}
\end{table}
%%%%%%%%%%%%%%%%%%%%%%%%%%%%%%%%%%%%%%%%%%%%%%%%%%%%%%%%%%%%%%%%%%%%%%%%%%%%%%%
%%%%%%%%%%%%%%%%%%%%%%%%%%%%%%%%%%%%%%%%%%%%%%%%%%%%%%%%%%%%%%%%%%%%%%%%%%%%%%%
\begin{table}[t!]
    \centering
    \tablestyle{1.8pt}{1.05}
    \begin{tabular}{l|l|c|c}
        \specialcell{Stage-Transition \\Fusion} & Termination Fusion & Train Acc & Top-1 Acc \\
        \shline
        Max & Norm $\rightarrow$ Concat & 78.1 & 81.7 \\ 
        Concat  & Norm $\rightarrow$ Concat & 79.1 &  82.0  \\ 
        \hline
        2\x-MLP    & Norm $\rightarrow$ 2\x-MLP & 80.2 & 81.8 \\ 
        2\x-MLP + 0.2 dp   & Norm $\rightarrow$ 2\x-MLP $\rightarrow$ 0.5dp & 77.1 & 81.2 \\ 
        2\x-MLP    & Norm $\rightarrow$ 1-layer & 53.6 & 82.1 \\
        2\x-MLP    & Norm $\rightarrow$ 1-layer $\rightarrow$ 0.2dp & 64.0 & 82.4 \\ 
        \hline
        Norm $\rightarrow$ 2\x-MLP    & Norm $\rightarrow$ Concat  & 79.4 & 82.3 \\
        Norm $\rightarrow$ 2\x-MLP    & \specialcell{Norm $\rightarrow$ 1-layer $\rightarrow$ \\  0.2dp $\rightarrow$ Norm}  & 78.3 & 82.3 \\ 
        \hline
        4\x-MLP   & Norm $\rightarrow$ Concat & 80.4 & 82.3 \\ 
        2\x-MLP    & Concat $\rightarrow$Norm & 80.5 & 82.2 \\ 
        \rowcolor{baselinecolor}
        2\x-MLP    & Norm $\rightarrow$ Concat & 80.1 & 82.5 \\ 
    \end{tabular}
    \vspace{-6pt}
    \caption{\textbf{Lateral Fusion Strategies}. Residual streams $I_1$ and $I_2$ are fused in state-transition blocks (\sref{sec:method:mvit:transition}) as well as on termination (\sref{sec:method:vit:boundary}) before the network head. We find fusion strategy to play a key role for ReV-MViT performance. Rev-MViT-B uses a 2-layer MLP with 2\x\ hidden dimensions in stage-transition blocks (gray). Please see Section \ref{sec:ablations} for details.}
    \label{tab:ablation:fusion}
    \vspace{-10pt}
\end{table}

%%%%%%%%%%%%%%%%%%%%%%%%%%%%%%%%%%%%%%%%%%%%%%%%%%%%%%%%%%%%%%%%%%%%%%%%%%%%%%%
\paragraph{Lateral Fusion Strategies.}
The stage-transition blocks employ residual stream fusion blocks for mixing information between $I_1$ and $I_2$ (\sref{sec:method:mvit:transition}). We explore several fusion strategies in Table~\ref{tab:ablation:fusion} using a combination of: (A) \textbf{$n$\x-MLP}: Two layer perceptrons with $n$ times the hidden dimension and GELU activations. (B) \textbf{$0.n$ dp}: $n\times10$ percent dropout on output activations. (C) Simple operators such as channel-wise maximum of $I_1$ and $I_2$ activations, and channel-wise concatenation of tensor. 

Lateral connections in stage-transition stages allows effective information mixing between the residual streams and hence increases network capacity. For example, compared to concatenation, 2\x-MLP increases to training accuracy by $1\%$ and also the top-1 performance by $0.5$\%. However an even heavier strategy, such as 4\x-MLP  widens the generalization gap and promotes over-fitting. Note that the training accuracy is often lower than the top-1 performance because of training data augmentations.  

\paragraph{Re-configuring residual connections.} 
As discussed in \sref{sec:method:vit:drift}, the reversible vision transformer design removes the skip connections that are commonly used inside the Attention and MLP blocks (Figure \ref{fig:rev_all}). Specifically, for all of the reversible blocks in Rev-MViT and Rev-ViT, the inputs $I_1$ and $I_2$ do not have residual connections that allow residual signal propagation by directly skipping their respective functions (MLP for $I_2$ and Attention for $I_1$). Instead their residuals are performed via the other residual stream operating through the reversible transform $T$ (\sref{sec:method:transform}). 

We ablate this design choice for the ViT architecture in Figure \ref{fig:ablation:residual}. We vary the model depth without changing any other model dimensions and observe the performance of the two reversible variants \ie, with and without internal skip connections. We note that while the na\"ive version with internal skip connections trains stably for shallow models, for deeper models the accuracy drops significantly. On the other hand, Rev-MViT scales well with depth, just as accurate as the other variant at shallower depths but significantly better with deeper models. 

\paragraph{Effect of learning without caching activations.} Figure \ref{fig:ablation:residual} also compares the image classification performance of the Rev-ViT-B architecture trained with and without caching activations. This allows us to disentangle the effect of the proposed residual configurations necessary for reversible vision transformer from any artefacts that might result from learning without caching activations. However, for all depths we notice the Rev-ViT-B performance trained without caching activations to closely track the performance of the same architecture trained with caching. The slight difference at depth 12 results might stem from the training recipe being adapted for the actual Rev-ViT-B architecture trained without activations. 

\paragraph{Model size and Input Resolution \vs Throughput.} Figure \ref{fig:ablation:throughput} shows the training throughput comparisons for different models sizes at $224$ and $320$ input resolutions. We note that while for smaller models such as, MViT-B with a depth 12 layers, Rev-MViT-B has a slightly smaller training throughput (98.5 vs. 86.0), the additional re-computation burden of intermediate activations is easily overcome at both higher resolution training as well as for deeper models. In particular, at 224 resolution, the 24-layer and 48-layer Rev-MViT models have similar throughput as the MViT models increasing upto to \textbf{2.1}\x\ higher throughput at 384 resolution and \textbf{2.3}\x\ higher throughput at 384 resolution for the 80 layer depth models. Further, the rate of memory increase for deeper models is much lower for reversible variants than vanilla networks, allowing scaling to much deeper models without any additional training infrastructure burden or memory requirement like with gradient checkpointing or model parallelism. 

\paragraph{Maximum batch-size.} We benchmark the maximum possible batch size for Rev-ViT Base (B), Large (L) and Huge (H) and their non-reversible counterparts in  Fig.\ref{fig:ablation:bscdepth}. We extrapolate the trend (denoted by \mediumdashes) to larger models by scaling ViT-L and ViT-H in depth (keeping other model dimensions constant) and benchmark the maximum batch size for their reversible counterparts.

\paragraph{Model size vs. GPU memory footprint.} Figure \ref{fig:teaser} plots the GPU Memory footprint for both Rev-ViT and Rev-MViT family of models as well as for several other prior networks such as MViT~\cite{MViT}, ViT~\cite{ViT}, ResNets and RegNetY~\cite{ilija_2020}. We note that at fixed GFLOPs, reversible variants are extremely memory efficient going upto $\textbf{4.5}$\x\ for MViT and $\textbf{15.5}$\x\ for ViT surpassing prior convolutional variants by orders of magnitude. 

\section{Conclusion}
\noindent We present Reversible Vision Transformers, memory-efficient reversible architectural adaptations of ViT and MViT models. We benchmark across several tasks, such as image classification, object detection and video classification and across several metrics, such as model complexity, throughput, accuracy and memory usage. Given any specification, our {Rev-ViT} and {Rev-MViT} match the accuracy of non-reversible variants at a tiny fraction of the memory cost while maintaining similar training throughput for smaller models and up to 2.3\x\ higher throughput for larger models. Specifically, we observe that the Rev-ViT and the Rev-MViT models achieve upto 15.5\x~and 4.5\x~lighter memory footprint than ViT and MViT models respectively.
% In future work, we aim to use the proposed reversible model as key design space element for deeper and memory-efficient visual recognition architectures.   
\appendix

\section*{Follow-Up Work}
Chen~\etal~\cite{zhao2022re} apply the proposed Reversible Vision Transformer architecture design for Reversible Swin Transformers and apply the model for memory-efficient temporal action localization. Temporal action localization involves detecting precise temporal frame positions for the start and end boundaries of an action and hence needs to be performed at on a densely sampled video. Dense frame sampling causes GPU memory overheads during training that prohibit finetuning the backbone end-to-end on the temporal action localization task (TAL). Reversible backbone alleviate the memory overhead and allow efficient end-to-end TAL training, thereby providing significant localization performance boost. 

Concurrently, \cite{parallelvit} proposes a training procedure for reversible transformers that allows speeding up training while ensuring exact replication of the original computation. In particular, \cite{parallelvit} proposes to stagger the activation re-computation one transformer block ahead of the gradient computation using the recomputed activations of the previous block. This allows the activations to be available for gradient calculation of the next block, as soon as the previous block finishes. Hence, the gradient calculation step does not need to wait for activation recomputation, \textit{effectively} hiding the latency of the burden of re-computation behind latency of graident calculations, thus effectively speeding up training. This requires maintaining separate cuda work-streams that process the above two steps asynchronously. Depending on the hardware and computation size, there can be significant speedups from such operator parallelization.       

\section*{Acknowledgements}
The authors would like to thank Harshayu Girase for help with benchmarking models, Amir Gholami, Ajay Jain and Nikita Kiatev for helpful research discussions and reference suggestions, Ajay Jain, Matthew Tancik and Hang Gao for writing discussions and Shubh Gupta, Suzie Petryk, Hang Gao, Abhinav Agarwal, Medhini Narasimhan and Amur Ghosh for proofreading the manuscript.

\section*{Appendix}
 
\section{Architecture Details} 
\label{sec:details}
\paragraph{Reversible Vision Transformers} Table \ref{tab:arch_instances} shows the architectures for all the Reversible Vision Transformer Models. All models closely follow the original ViT architectures~\cite{ViT} in matched performance, parameters, FLOPs and much lower memory footprint (Table ~\ref{tab:sota:in1k}). Output sizes denote the tensor shapes of the two residual streams at the end of each reversible Vision Transformer block. Note that even though the intermediate activations are twice the non-reversible variant, the actual memory needed is much lower because of memory reuse in reversible training. Further, the FLOPs are matched since each layer is performed only one of the two streams. 
\paragraph{Reversible Multiscale Vision Transformers} Table \ref{tab:revmvit_arch} shows the architecture for the Rev-MViT-B model for image classification. The backbone is made-up of two stages -- Stage-transition blocks that increase the channel capacity and down-sample the resolution and the reversible Stage-preserving blocks that perform the majority of computation without changing feature dimensions. Similar to Rev-ViT, the output sizes of both the streams are denoted. Fusion blocks operate on $Y_1$ and $Y_2$ together, hence operate with computationally light operations (Table \ref{tab:ablation:fusion}).

\section{Training Settings} 
\label{sec:training_settings}
\paragraph{ImageNet.} Table \ref{tab:recipe} shows the training recipes for ViT-L and Rev-ViT-L models presented in Table ~\ref{tab:sota:in1k}. Note that ViT-L is quite heavy with 61.6 GFLOPs and hence we adopt a shorter 200 epochs recipe for faster experiment cycle for developing Rev-ViT-L. Smaller ViT models -- ViT-S and ViT-B -- are trained according to the Data efficient transformers~\cite{deit} and are all trained for 300 epochs. Hence, the accuracy difference between ViT-L which achieves 81.5\% while ViT-B achieves 81.8\% overall. MViT-B model follows the 300 epochs recipe as well proposed in \cite{MViT}. 
\paragraph{Kinetics-400 \& Kinetics-600.} We follow the recipes proposed in \cite{MViT} to train the Rev-MViT-B architecture (Table~\ref{tab:revmvit_arch}) following crucial modifications shown in Table \ref{tab:ablation:regularization}. 

\paragraph{MS-COCO.}
For object detection experiments, we adopt the  Mask R-CNN~\cite{He2017} object detection framework in Detectron2~\cite{wu2019detectron2}. We follow the same training settings from~\cite{liu2021swin}, AdamW optimizer~\cite{loshchilov2018fixing} ($\beta_1, \beta_2 = 0.9, 0.999$, base learning rate $\expnum{1.6}{-4}$ for base size of 64, and weight decay of 0.1), and 3x schedule (36 epochs). The drop path rate is set as $0.4$. We use PyTorch's automatic mixed precision during training.

\begin{table}[t!]
	\footnotesize
	\tiny 
	\centering
	\captionsetup[subffloat]{justification=centering}
	\subfloat[\textbf{Rev-ViT}-S  with \textbf{4.6}G FLOPs, \textbf{22}M param, \textbf{8.8}MB/img memory, and \textbf{79.9}\% top-1 accuracy.
	\label{tab:arch_vitb}]{
		\tablestyle{1pt}{1.05}
		\tablestyle{1pt}{1.08}  \scriptsize 
		\begin{tabular}{c|c|c}
			stage & operators & output sizes  \\
			\toprule
			\multirow{1}{*}{data} &     &  \outsizesRawAdapt{\tcolor{8}}{\xycolor{224}}{\xycolor{224}}{1}   \\
			\midrule
			
			\multirow{2}{*}{patch} & \multicolumn{1}{c|}{1\x16\x16, {384}} &   
			\outsizesRawDAdapt{\wcolor{$384$}}{\tcolor{8}}{\xycolor{14}}{\xycolor{14}}{2}    \\
			& stride 1\x16\x16   \\
			\midrule
			\multirow{2}{*}{rev}  & \blockatt{384}{{1536}}{12} &
			\outputstack{\wcolor{$384$}}{\xycolor{14}}{\xycolor{14}} \\ 
			&  & \\
			\bottomrule
		\end{tabular}
	}
	\hspace{6pt}
	\subfloat[\textbf{Rev-ViT}-B  with \textbf{17.6}G FLOPs, \textbf{87}M param, \textbf{17}MB/img memory, and \textbf{81.8}\% top-1 accuracy.	
	\label{tab:arch_m}]{
		\tablestyle{0.5pt}{1.05}
		\tablestyle{1pt}{1.08}  \scriptsize 
		\begin{tabular}{c|c|c}
			stage & operators & output sizes  \\
			\toprule
			\multirow{1}{*}{data} &     &  \outsizesRawAdapt{\tcolor{8}}{\xycolor{224}}{\xycolor{224}}{1}   \\
			\midrule
   			
			\multirow{2}{*}{patch} & \multicolumn{1}{c|}{1\x16\x16, {768}} &    \outsizesRawDAdapt{\wcolor{$768$}}{\tcolor{8}}{\xycolor{14}}{\xycolor{14}}{2}    \\
			& stride 1\x16\x16   \\
			\midrule
			\multirow{2}{*}{rev}  & \blockatt{768}{{3072}}{12} &
			\outputstack{\wcolor{$768$}}{\xycolor{14}}{\xycolor{14}}
			     \\ [8pt]
			\bottomrule
		\end{tabular}
	}		
	\hspace{6pt}
	\subfloat[\textbf{Rev-ViT}-L  with \textbf{61.6}G FLOPs, \textbf{305}M param, \textbf{22.6}MB/img memory, and \textbf{81.4}\% top-1 accuracy.
	\label{tab:arch_s}]{
		\tablestyle{0.5pt}{1.05}
		\tablestyle{1pt}{1.08}  \scriptsize 
		\begin{tabular}{c|c|c}
			stage & operators & output sizes  \\
			\toprule
			\multirow{1}{*}{data} &     &  \outsizesRawAdapt{\tcolor{8}}{\xycolor{224}}{\xycolor{224}}{1}   \\
			%			&  &  \\
			\midrule
			
			\multirow{2}{*}{patch} & \multicolumn{1}{c|}{1\x16\x16, {1024}} &    \outsizesRawDAdapt{\wcolor{$1024$}}{\tcolor{8}}{\xycolor{14}}{\xycolor{14}}{2}    \\
			& stride 1\x16\x16   \\
			\midrule
			\multirow{2}{*}{rev}  & \blockatt{1024}{{4096}}{24} &
			\outputstack{\wcolor{$1024$}}{\xycolor{14}}{\xycolor{14}}
			\\ [8pt]
			\bottomrule
		\end{tabular}
	}
	\caption{\textbf{Reversible Vision Transformer Architectures}: Rev-ViT are reversible adaption of ViT with exactly matched FLOPs, parameters and accuracy under identical conditions but with much lower GPU memory footprints. 
	}
	\label{tab:arch_instances}
\end{table}
\begin{table}[t!]
	\centering
	{
		\tablestyle{0.5pt}{1.05}
		\tablestyle{1pt}{1.08}  \scriptsize 
		\begin{tabular}{c|c|c}
			stage & operators & output sizes \\
			\toprule
			\multirow{1}{*}{data} &   &  \outsizesRawAdapt{\tcolor{\textbf{16}}}{\xycolor{224}}{\xycolor{224}}{1}   \\
			\midrule
			
			\multirow{2}{*}{cubification} & \multicolumn{1}{c|}{7\x7, {96}} &    \outsizesRawDAdapt{\wcolor{$96$}}{\tcolor{\textbf{8}}}{\xycolor{56}}{\xycolor{56}}{2}    \\
			& stride 4\x4   \\
			\midrule
			\multirow{2}{*}{Stage-Preserving}  & \blockatta{96}{{384}}{1} &
			\outputstack{\wcolor{$96$}}{\xycolor{56}}{\xycolor{56}}
                \\
			&  & \\
			\midrule
			\multirow{3}{*}{Stage-Transition}  & \blockattfus{192}{{768}}{1} & \outsizesRawDAdapt{\wcolor{$192$}}{\tcolor{\textbf{8}}}{\xycolor{28}}{\xycolor{28}}{3}  \\ [18pt]
			\multirow{2}{*}{Stage-Preserving}  & \blockatta{192}{{768}}{1} &
			\outputstack{\wcolor{$192$}}{\xycolor{28}}{\xycolor{28}}
			\\
			&  & \\
			\midrule
			\multirow{3}{*}{Stage-Transition}  & \blockattfus{384}{{1536}}{1} & \outsizesRawDAdapt{\wcolor{$384$}}{\tcolor{\textbf{8}}}{\xycolor{14}}{\xycolor{14}}{3}  \\ [18pt]
			\multirow{2}{*}{Stage-Preserving}  & \blockatta{384}{{1536}}{10} &
			\outputstack{\wcolor{$384$}}{\xycolor{14}}{\xycolor{14}}
			\\
			&  & \\
			\midrule
			\multirow{3}{*}{Stage-Transition}  & \blockattfus{768}{{3072}}{1} & \outsizesRawDAdapt{\wcolor{$768$}}{\tcolor{\textbf{8}}}{\xycolor{7}}{\xycolor{7}}{3}  \\ [18pt]
			\multirow{2}{*}{Stage-Preserving}  & \blockatta{768}{{3072}}{1} &
			\outputstack{\wcolor{$768$}}{\xycolor{7}}{\xycolor{7}}
			  \\
			&  & \\
			\hline
		\end{tabular}
		}
		\caption{\textbf{Rev-MViT}-B with \textbf{8.7}G FLOPs, \textbf{39}M param, \textbf{66.8}MB/img memory, and \textbf{82.5}\% top-1 accuracy is reversible adaption of MViT-B architecture~\cite{ViT}. }
		\label{tab:revmvit_arch}
\end{table}
\begin{table}[t!]
    \centering
    \tablestyle{1.8pt}{1.05}
    \begin{tabular}{l|cc}
        \multirow{1}{*}{Training Hyperparameter} & ViT-B  & Rev-ViT-B \\
        \shline
        Learning Rate                                             & 1e-4 & 7e-5 \\
        Random augment Repeats (N)    & 1 & 2\\
        Random augment Magnitude (M)    & 9 & 7\\
        Optimizer Momentum   & (0.9, 0.95) & (0.9, 0.999) \\ 
        \hline
        Weight Decay                                 & 0.3 & 0.3\\
        Batch Size & 4096 & 4096 \\ 
        Epochs  & 200 & 200 \\
        Label Smoothing      & 0.1 & 0.1 \\
        Drop Path Rate       & 0.2 & 0.2 \\
        Mixup                & 0.8 & 0.8 \\
        Cutmix               & 1.0 & 1.0 \\
    \end{tabular}
    \caption{\textbf{Training Recipe for ViT-L and Rev-ViT-L
    }}
    \label{tab:recipe}
\end{table}

%%%%%%%%% REFERENCES
{\small
\bibliographystyle{ieee_fullname}
\bibliography{mvitl4}
}

\end{document}